\documentclass[10pt, a4paper]{article}

\usepackage{etoolbox}
\patchcmd{\quote}{\rightmargin}{\leftmargin}{}{}
\patchcmd{\quote}{\list}{\list{}{\leftmargin=0pt \rightmargin=0pt}}{}{}

\usepackage[final]{lrec2026}
\usepackage{graphicx}
\usepackage{todonotes}
\usepackage{booktabs}
\usepackage{linguex} % package for example formating
\usepackage{multirow}

\usepackage{tikz}
\usetikzlibrary{trees,positioning,shapes,shadows,arrows}
\tikzset{
  basic/.style  = {draw, text width=2cm, drop shadow, font=\sffamily, rectangle},
  root/.style   = {basic, rounded corners=2pt, thin, align=center, fill=white, text width=3.8cm},
  level-2/.style = {basic, rounded corners=6pt, thin,align=center, fill=white, text width=3cm},
  level-3/.style = {basic, thin, align=center, fill=white, text width=2.3cm},
  level-4/.style = {basic, thin, align=center, fill=white, text width=2.5cm}
}

\title{GUMBridge: a Corpus for Varieties of Bridging Anaphora}

\name{Lauren Levine, Amir Zeldes} 

\address{Georgetown University, Department of Linguistics \\
\{lel76, amir.zeldes\}@georgetown.edu\\}

\abstract{
Bridging is an anaphoric phenomenon where the referent of an entity in a discourse is dependent on a previous, non-identical entity for interpretation, such as in "There is \underline{a house}. \textbf{The door} is red," where the door is specifically understood to be the door of the aforementioned house. While there are several existing resources in English for bridging anaphora, most are small, provide limited coverage of the phenomenon, and/or provide limited genre coverage. In this paper, we introduce GUMBridge, a new resource for bridging, which includes 24 diverse genres of English, providing both broad coverage for the phenomenon and granular annotations for the subtype categorization of bridging varieties. 
We also present an evaluation of annotation quality and report on baseline performance using open and closed source contemporary LLMs on three tasks underlying our data,
showing that bridging resolution and subtype classification remain difficult NLP tasks in the age of LLMs. 
 \\ \newline  }

\begin{document}

\maketitleabstract

\section{Introduction}

% What is bridging, overview of the resource we are presenting and why it is necessary, evaluation of LLMs using the resource (demonstrating that the task is till very challenging in the era of LLMs), and error analysis comparing human aanotation to LLM performance.

When an entity is mentioned in a discourse, it is generally either already familiar to the participants or it is new to them. However, there are also instances where an entity is newly introduced to the discourse but the identity of its referent is still inferable to participants. When the referent is inferable specifically due to the presence of a previous, non-identical entity in the discourse, we consider it to be a case of "bridging anaphora". Consider the following sentence:

\ex.
While dancing \underline{ballet} takes dedication and requires serious training, you can learn \textbf{the basics} to prepare yourself for \textbf{further study}.\footnote{Bridging anaphora are marked in bold face, and their associative antecedents are underlined.}
\label{ex:opening}

In \ref{ex:opening} above, the bold entities "the basics" and "further study" are newly introduced, but it is inferable that they specifically refer to "the basics of ballet" and "further study of ballet". These entities are bridging anaphora, and "ballet" is their associative antecedent. "Bridging" occurs when a discourse participant constructs an implicature from the entity they are currently processing back to an antecedent entity \citep{clark-1975-bridging}. The associative relations between entities that give rise to bridging links can manifest in a variety of different ways in a discourse, including part-whole relations (\underline{a house} $\rightarrow$ \textbf{the door}), relative adjectives (\underline{a dog} $\rightarrow$ \textbf{a larger dog}), and prototypical associations (\underline{a library} $\rightarrow$ \textbf{the books}). The correct interpretation of such relations is essential for a variety of NLP tasks, including question-answering, summarization and more, and it is an open question to what extent current LLMs can resolve them (see Section \ref{sec:LLM}). 

% relational nouns (\underline{parent} $\rightarrow$ \textbf{child})

Although bridging is relatively understudied when compared with other entity related tasks such as named entity recognition (NER) and coreference resolution, bridging resolution -- the automatic identification of bridging anaphora and resolution back to their associative antecedents -- has gained attention in recent years \citep{kobayashi-etal-2023-pairspanbert, hou-2020-bridging}. Additionally, efforts have been made to include bridging anaphora in recent shared task settings \citep{khosla-etal-2021-codi, yu-etal-2022-codi}. However, despite some attempts at harmonization \citep{levine-zeldes-2024-unifying}, the current resource landscape for bridging remains disparate, containing a number of different annotation formalisms with varying definitions of bridging and annotation schemes for subcategorization  \citep{kobayashi-ng-2020-bridging}. 

Since its introduction, bridging has been studied from a variety of theoretical perspectives (e.g., \citealp{hawkins1978definiteness}; \citealp{prince1981toward}; \citealp{asher1998bridging}; \citealp{baumann2012referential}), and linguistic resources have been constructed for various languages, including German (e.g., GRAIN \citeplanguageresource{schweitzer-etal-2018-german} and DIRNDL \citeplanguageresource{Eckart2012}), Polish (e.g., PCC \citeplanguageresource{ogrodniczuk-zawislawska-2016-bridging}), and Czech (e.g., PDT \citeplanguageresource{nedoluzhko-etal-2009-coding}). In English, there are several influential resources which have been used for the evaluation of bridging resolution: ISNotes \citeplanguageresource{markert-etal-2012-collective}, BASHI \citeplanguageresource{rosiger-2018-bashi}, and ARRAU RST \citeplanguageresource{poesio-artstein-2008-anaphoric, Uryupina2019AnnotatingAB}. ARRAU annotates instances of bridging by identifying entity pairs that establish cohesion, using a set of predefined semantic relations, while the bridging annotations of ISNotes and BASHI are based on the information status of entities, regarding bridging as a type
of mediated information. Such differences complicate setting up a standard benchmark for evaluation, as it is difficult to make a meaningful comparison of results of bridging resolution systems on different resources. Additionally, these resources are mostly small, severely lacking in genre diversity, and/or provide coverage of the phenomenon that is limited in various ways. 

% (ISNotes and BASHI each containing less than 1k instances of bridging)
% (all mostly composed of Wall Street Journal (WSJ) news data from more than 30 years ago)

% (including lexical bridging, see below)

In order to bridge this gap in the landscape of bridging resources, we present GUMBridge, a new resource for bridging, which contains \textbf{5.7k bridging instances} over \textbf{291k tokens} of English, covering \textbf{24 diverse genres}. GUMBridge is constructed on top of GUM v12, an existing multi-genre corpus of English \citeplanguageresource{Zeldes2017}, and it aims to unite aspects of current bridging formalisms for English. GUMBridge provides broad coverage for the phenomenon of bridging, using rigorously defined information status based criteria to identify instances of bridging (similar to ISNotes and BASHI), which does not place structural limitations on the manner in which bridging can manifest in a discourse. Additionally, GUMBridge provides annotations for  categorizing subtypes of bridging (as in ARRAU), introducing a new schema with 10 subtypes divided into 3 main categories: \textsc{comparison}, \textsc{entity}, and \textsc{set} (with an additional \textsc{other} category). GUMBridge also allows for the annotation of multiple subtypes on a single instance of bridging, which has not been possible in any previous English resource. Leveraging this data, we provide a baseline evaluation of three  LLMs on the task of bridging resolution, with the additional task of subtype categorization. Our results show that bridging resolution remains a difficult task, even for state-of-the-art large language models. 
%A brief cross-genre analysis of our results also shows that spoken data is more difficult for the LLM than written data.

\section{Background}
\label{sec:background}

% Detailed overview on the bridging research and resource landscape, motivating the need for a new resource that fills in the gaps

% 3 main resources
% information status vs cohesion and pre-defined semantic relations

As noted above, the current landscape of bridging resources is heterogeneous, with resources covering the scope of the phenomenon differently. In this section, we discuss the relevant theoretical divides and gaps in coverage in the landscape of English bridging resources. We give details on the ISNotes, BASHI, and ARRAU corpora, which are commonly used for the evaluation of bridging resolution systems \cite{yu-etal-2022-codi, kobayashi-etal-2023-pairspanbert}. We also give details on the original GUM corpus, the resource on top of which GUMBridge is constructed.

\begin{table}[]
\centering
\resizebox{\columnwidth}{!}{%
\begin{tabular}{lccc}
\toprule
 & \textbf{Tokens} & \textbf{\begin{tabular}[c]{@{}c@{}}Bridging\\ Instances\end{tabular}} & \textbf{\begin{tabular}[c]{@{}c@{}}Bridging per \\ 1k Tokens\end{tabular}} \\ \midrule
\textbf{ISNotes} & 40k & 916 & 22.7 \\
\textbf{BASHI} & 58k & 459 & 7.9 \\
\textbf{GUM v11} & 268k & 2.2k & 8.2 \\ \midrule
\textbf{ARRAU} & \multicolumn{1}{l}{} & \multicolumn{1}{l}{} & \multicolumn{1}{l}{} \\
\hspace{5mm}RST & 229k & 3.8k & 16.5 \\
\hspace{5mm}GNOME & 21k & 692 & 32.2 \\
\hspace{5mm}PEAR & 14k & 333 & 23.7 \\
\hspace{5mm}TRAINS & 84k & 710 & 8.5 \\ \midrule
\hspace{5mm}Total & 348k & 5.5k & 15.8 \\ \midrule
\textbf{\begin{tabular}[l]{@{}l@{}}GUMBridge v1\end{tabular}} & \multicolumn{1}{l}{} & \multicolumn{1}{l}{} & \multicolumn{1}{l}{} \\
\hspace{5mm}Train & 213k & 4k & 18.9 \\
\hspace{5mm}Dev & 30k & 732 & 24.5 \\
\hspace{5mm}Test & 30k & 562 & 18.6 \\ 
\hspace{5mm}Test2 & 18k & 379 & 21.2 \\ \midrule
\hspace{5mm}Total & 291k & 5.7k & 19.6 \\ \bottomrule 
\end{tabular}%
}
\caption{Comparison of the size of several bridging resources for English.}
\label{tab:corpora_stats}
\end{table}

%\begin{table}[h!tb]
%\centering
%\resizebox{\columnwidth}{!}{%
%\begin{tabular}{lccc}
%\hline
% & Tokens & \begin{tabular}[c]{@{}c@{}}Bridging\\ Instances\end{tabular} & \begin{tabular}[c]{@{}c@{}}Bridging per \\ 1k Tokens\end{tabular} \\ \hline
%ARRAU RST & 229k & 3.8k & 16.5 \\
%ISNotes & 40k & 663 & 16.6 \\
%BASHI & 58k & 459 & 7.9 \\
%GUM v11 & 268k & 2.2k & 8.2 \\
%GUMBridge v1 & 127k & 2.3k & 17.1 \\ \hline
%\end{tabular}%
%}
%\caption{Comparison of several bridging resources for English.}
%\label{tab:corpora_stats}
%\end{table}

\paragraph{Information Status \& Bridging} The information status (IS) of an entity refers to the extent to which that entity is familiar to the participants of a discourse \citep{nissim-etal-2004-annotation}. Broadly speaking, an entity can either be "New" information, where the entity is not yet known to the participant, or it can be "Given" information, where the entity is recognized by the participant. If an entity is recognized, this can be for various reasons: (1) the entity may have been already introduced in
the discourse (coreference), (2) the entity may be inferable via generic/world knowledge or deixis (pointing to something in context), or (3) the entity may be inferable from a previous, non-identical entity in the discourse, which is the case of bridging anaphora. Although entities understood through bridging and generic/world knowledge are both considered "Accessible" (as they are recognized by discourse participants when first introduced to the discourse), only instances of bridging depend on the presence of an associative antecedent for comprehension.

%In early work relating to bridging, \citet{prince1981toward} discusses the concept of \textit{Inferables}, entities for which the reader/hearer can interpret their referent via a plausible inference from an existing entity in the discourse. This framing acts to center information status as the key component in identifying anaphoric bridging relations. Such an information status informed view of the criteria for an instance of bridging is adopted in ISNotes and BASHI, which consider bridging to be a type of mediated information. We also adopt such a view in the construction of GUMBridge.

\paragraph{Lexical Bridging vs. Referential Bridging}
The distinction between lexical bridging and referential bridging is a theoretical divide that stems from the decision of most corpora to classify bridging as an anaphoric phenomenon based on information status. When introducing the BASHI corpus, \citet{rosiger-2018-bashi} coined the terms for this distinction as a means of differentiating the types of bridging annotations present in different corpora. Referential bridging refers to truly anaphoric instance of bridging where the bridging anaphor requires an antecedent to be interpretable, as in \ref{ex:referential}: 

\ex. She likes \underline{the house} because \textbf{the windows} are large.
\label{ex:referential}

Lexical bridging refers to lexical semantic relations between pairs of entities, such as part-whole or set-member relations, which may or may not be anaphoric, as in \ref{ex:lexical} where the antecedent is not strictly necessary for interpretation:

\ex. I went to \underline{the United States} last month. My first stop was \textbf{Washington, DC}.
\label{ex:lexical}

While ARRAU includes instances of both lexical bridging and referential bridging, ISNotes and BASHI both limit themselves to including only referential bridging. From an information status based perspective, instances of lexical bridging are not considered bridging at all if they are not also anaphoric in nature. As such, resources such as ARRAU, which include various types of lexical bridging, have received criticism for being over-annotated \cite{roesiger-etal-2018-bridging}.

\paragraph{Corpora} The following are short descriptions of the resources which have most strongly informed the construction of GUMBridge. A comparison of the size of these resources in tokens and instances of bridging is shown in Table~\ref{tab:corpora_stats}.

\textbf{ISNotes }is a corpus for annotations of fine-grained information status, and it is composed of 50 Wall Street Journal (WSJ) documents (40k tokens). Considering bridging to be a type of mediated information, it contains 663 instances of bridging in the \texttt{mediated/bridging} subcategory and an additional 253 instances of comparative anaphora (commonly considered to be a subvariety of bridging) in the \texttt{mediated/comparison} subcategory (e.g.~cases like \textbf{a larger dog} above), for a total of 916 instances.

%ISNotes is a corpus for annotations of fine-grained information status, and it is composed of 50 Wall Street Journal (WSJ) documents taken from the OntoNotes corpus \citeplanguageresource{weischedel2011ontonotes}. ISNotes distinguishes three main categories of IS: \texttt{New}, \texttt{Old}, and \texttt{Mediated}. There are six subcategories within the \texttt{Mediated} category, one of which is \texttt{bridging}. There are 663 instances of bridging in the \texttt{mediated/bridging} subcategory, approximately 16.6 bridging instances per 1k tokens. Additionally, there are an 253 instances of comparative anaphora in the \texttt{mediated/comparison} subcategory, which is commonly considered to be a subvariety of bridging. These categories are both types of referential bridging. The Cohen's $\kappa$ reported by \citet{markert-etal-2012-collective} shows that \texttt{mediated/bridging} is more difficult than other IS categories, ranging \textasciitilde0.6-0.7 for \texttt{mediated/bridging}, and reaching \textasciitilde0.8 for \texttt{mediated/comparison}.

\textbf{BASHI} is a corpus that specifically targets the annotation of bridging anaphora, and it also covers a set of 50 WSJ documents (58k tokens, different documents than ISNotes). As noted above, \citet{rosiger-2018-bashi} introduces the distinction between referential and lexical bridging, and BASHI explicitly includes only instances of referential bridging in its construction. It contains just 459 annotated bridging pairs.

%, corresponding to approximately 7.9 instances per 1,000 tokens.  The annotated bridging instances fall into three subtypes: definite, indefinite, and comparative anaphora. Inter-annotator agreement was measured for each subtype as well as overall. The overall agreement for identifying bridging instances is 59.3\%, with higher agreement for comparative anaphora (71.4\%), and lower agreement for definite (63.8\%) and indefinite (42.3\%) instances.

\textbf{ARRAU} is composed of four subcorpora, covering various genres: ARRAU RST (WSJ news), ARRAU PEAR (spoken narrative), ARRAU TRAINS (dialogues), and ARRAU GNOME (medical/art history). However, the annotations for the subcorpora are not consistent, and the largest subcorpus, ARRAU RST (229k tokens) is the only one that has been used in evaluation of bridging resolution systems. It includes 3,777 of the 5,512 bridging annotations in ARRAU. ARRAU’s bridging annotation links related mentions that establish entity coherence through non-identity relations. Because this definition is intentionally broad, annotation is constrained to a fixed set of semantic relations.

%The ARRAU corpus is a broad effort, aiming to capture a wide range of anaphoric phenomena, including bridging. 

%The corpus distinguishes nine bridging subtypes, including \texttt{possession}, \texttt{element–set}, \texttt{subset–set}, and `other' (along with their inverse relations), plus an additional \texttt{under-specified} category. The annotation scheme for bridging in ARRAU extends the framework developed in the GNOME project \citeplanguageresource{poesio-2004-discourse}. Annotators in GNOME achieved high agreement on subtype labeling at 95.2\%, but only 22\% recall when it came to identifying bridging instances. As previously noted, ARRAU contains instances of both lexical bridging and referential bridging. 

\textbf{GUM} (291k tokens, v12) is a resource that covers a variety of linguistic phenomena for 16 diverse genres of English: academic writing, biographies, courtroom transcripts, essays, fiction, how-to guides, interviews, letters, news, online forum discussions, podcasts, political speeches, spontaneous face to face conversations, textbooks, travel guides, and vlogs. The extended test set of the corpus, GENTLE \citeplanguageresource{aoyama-etal-2023-gentle}, is included in GUM v12 as the `test2' partition, and contains an additional 8 genres: dictionary entries, live esports commentary, legal documents, medical notes, poetry, mathematical proofs, course syllabuses, and threat letters. 

Version 11 of the GUM corpus had a total of 2,192 bridging instances (both lexical and referential bridging). Although GUM v11 already contained annotations for bridging, the annotation guidelines relied upon were minimal \citep{gum-wiki}, and GUMBridge aims is to broaden the coverage and robustness of the bridging annotations for GUM v12. In addition to the standalone data release accompanying this paper, the annotations of GUMBridge are included in the GUM v12 release,\footnote{https://github.com/amir-zeldes/gum} and GUMBridge style annotations will continue to be included in future GUM releases. 

% , broadly defining bridging as any underspecified, newly introduced entity whose identity is inferable via a non-identical antecedent entity
% corresponding to approximately 8.2 instances per 1k tokens.

\paragraph{Gaps in the Resource Landscape}

% and it is diffult to put the resources together because they are constructed with different formalism

From the descriptions of the English bridging resources above, we can see that no single one of them can be considered a reasonable "reference" corpus for bridging in English. First, the number of bridging instances present in existing resources is relatively low, as we can see in Table~\ref{tab:corpora_stats}. The resources that exclusively contain referential bridging are quite small, with ISNotes and BASHI containing only 916 and 459 instances of bridging respectively. And while ARRAU contains more instances of bridging (5,512), it is likely to be "over annotated" for studies only concerned with referential bridging, and it is not trivial to separate referential and lexical bridging in the data. Additionally, we can see in Table~\ref{tab:corpora_stats} that the frequency of bridging instances annotated in existing resources has been quite variable. ARRAU and ISNotes have approximately twice the frequency of bridging annotations per 1k tokens when compared with BASHI and GUM v11 (15.8 and 22.7 vs 7.9 and 8.2 respectively). This suggests that existing resources are likely under-annotated, and that broader coverage for bridging annotations is required. 

There is also a complete lack of genre diversity amongst the resources that have commonly been used for the evaluation of bridging resolution systems; ISNotes, BASHI, and ARRAU RST are all solely composed of WSJ data from more than 30 years ago, which predates basic modern expressions such as "cell phone" and "the internet". And while ARRAU provides some genre coverage in its sub-corpora, they are unbalanced, both in their size and in the density of their bridging annotations (see Table~\ref{tab:corpora_stats}). As a result of their internal inconsistency, they are not commonly used for evaluation. Additionally, ARRAU is the only resource that provides subtype annotations, meaning that more granular linguistic analysis is not possible for the resources focused on referential bridging. These gaps motivate the current paper's introduction of GUMBridge, a new resource which provides a substantial increase in both genre diversity and annotation coverage, as well as the addition of multi-subtype annotations for every instance of bridging.

% could add something here about the need to have the ability to annotate multiple subtypes which has been missing from previous resources

\section{Resource Description}

%\begin{table}[h!tb]
%\centering
%\resizebox{\columnwidth}{!}{%
%\begin{tabular}{lcccccc}
%\hline
% & Genres & Docs & Tokens & Mentions & \begin{tabular}[c]{@{}c@{}}Bridging\\ Instances\end{tabular} & \begin{tabular}[c]{@{}c@{}}Bridging per \\ 1k Tokens\end{tabular} \\ \hline
%Train & 16 & 64 & 67k & 19k & 1k & 16.3\\
%Dev & 16 & 32 & 30k & 8.4k & 601 & 20.1\\
%Test & 16 & 32 & 30k & 8.4k & 577 & 19.0\\ \hline
%Total & 16 & 128 & 127k & 35k & 2.3k & 17.1\\ \hline
%\end{tabular}%
%}
%\caption{GUMBridge corpus statistics.}
%\label{tab:gumbridge_stats}
%\end{table}

GUMBridge is a new resource for varieties of bridging anaphora, constructed on top of the existing GUM corpus. The bridging annotations for GUMBridge v1 are released with this paper under the Creative Commons Attribution (CC BY-NC-SA) version 4.0 license via our GitHub repository.\footnote{https://github.com/lauren-lizzy-levine/gumbridge} GUMBridge contains 5,698 instances of bridging over 291k tokens, corresponding to 19.6 instances per 1k tokens. As we can see in Table~\ref{tab:corpora_stats}, after ISNotes, this the highest overall coverage of the resources discussed. GUMBridge follows the train/dev/test/test2 splits established by GUM v12.
%, including all of the documents in GUM dev and test, and a portion of the documents in GUM train. 
Although we do not leverage our splits for model training in this paper, we provide official, genre balanced splits in order to promote consistency in future work.\footnote{\citet{kobayashi-ng-2020-bridging} note that when bridging resources lack official splits (like ISNotes and BASHI), evaluation is done using k-fold cross-validation, which hinders direct comparison of results.} The size of the splits in terms of tokens and bridging instances is shown in Table~\ref{tab:corpora_stats}. 

As previously noted, GUMBridge aims to unite desirable aspects of existing bridging resources: like ISNotes and BASHI, we take an information status informed approach to identifying instances of bridging, and like ARRAU, we provide subtype annotations for each instance of bridging, which allows for the analysis of different varieties of bridging anaphora. 
In the following subsections, we give details on GUMBridge's information status based annotation procedures for train/dev/test/test2 and its subcategorization schema for varieties of bridging. 

% information status basis
% leveraging existing GUM data which has existing annotations for entity spans/subtypes and coreference, annotation interface

\subsection{Annotation Procedure}

% Annotation of train, dev, test, 
The dev and test sets of GUMBridge were manually annotated by the authors of this paper and graduate students trained on the annotation guidelines below.\footnote{Full annotation guidelines included in supplementary materials and GitHub repository.} The annotation guidelines for GUMBridge were developed after careful review of the bridging definitions, annotation guidelines, and data of prominent English bridging resources (ISNotes, BASHI, ARRAU, and GUM), aiming to unite desirable aspects of each.
An initial pilot was conducted on the test set, which achieved low inter-annotator agreement on the recognition of bridging anaphora \citep{levine-zeldes-2025-subjectivity}. Following a careful analysis of the poor agreement results, which focused on the subjectivity of annotator judgments, the scheme for bridging subtype classification was revised and the annotation guidelines were significantly extended. Following the finalization of the scheme/guidelines, the authors of this paper revised the annotations on the test set.
An agreement study on the annotation of the dev set using the finalized annotation scheme and guidelines is included in Section~\ref{sec:IAA}. 

The train set and test2 of GUMBridge were manually annotated by the authors of this paper. While the bridging annotations in dev and test were done from scratch, the train and test2 documents were corrected and expanded from the original bridging annotations in GENTLE and  GUM v11.\footnote{The train documents added for GUM v12 were first annotated by students in a classroom setting using the GUMBridge guidelines (as part of GUM's annual expansion procedure), and then corrected by authors of this paper.} The original GUM corpus also already contains entity and coreference annotations following the same mention definitions used in this paper,\footnote{See \citet{Zeldes2022} for discussion of coreference guidelines in GUM, and an alternative version following OntoNotes guidelines \citeplanguageresource{WeischedelPradhanRamshawEtAl2012} in \citet{zhu-etal-2021-ontogum}} as well as information status annotations, i.e., "New", "Given", and "Accessible" (not including accessibility from instances of bridging), which were provided to the annotators of GUMBridge. Additionally, common structural triggers of bridging anaphora (e.g., 2 token definite entities, relative adjectives) were highlighted based on a review of the annotated dev and test data in order to help focus annotators and improve recall.  The annotation project was completed using an adapted version of the GitDox annotation interface \cite{Zhang2017GitDOXAL}.

\paragraph{Identifying Bridging Anaphora}

GUMBridge adopts an information status informed perspective to identifying instances of bridging anaphora. As explained in Section~\ref{sec:background}, the information status of an entity refers to the level of its ``Accessibility'' to the participants of a discourse. An entity is considered to be ``Accessible'' if it is newly mentioned yet is still comprehensible to the participants of a discourse. In the case of bridging, the ``Accessible'' status of the entity in question is specifically due to an inference being made from a previous, non-identical antecedent. This is in contrast with entities that are considered ``Accessible'' for other reasons, such as generics/world knowledge or situational context (e.g., unanchored deictic expressions). Bridging anaphora are not accessible by themselves, but dependent on previous entities for interpretation. Note this does not preclude named entities, accessible through world knowledge, from being bridging anaphora if they instantiate a bridging relation, though this is rare: using the existing Wikification annotations in GUM \cite{lin-zeldes-2021-wikigum} we find only 98 cases of bridging with linked named entities, as in \ref{ex:linking}.

\ex. (he) remained \underline{a lifelong Roman Catholic}, at a time when \textbf{the Church} was being persecuted\label{ex:linking}

Annotators receive a detailed explanation of this definition of bridging and accessibility and are instructed to apply the following three criteria when determining whether an entity constitutes a bridging anaphor:

\begin{enumerate}
    \item Is this entity Accessible (known/inferable at first mention) in the discourse?
    \item Is this entity Accessible because of a relationship to a previous, non-identical entity in the discourse? If so, identify the antecedent entity.
    \item Rule out the following scenarios which are \textbf{not} considered to be bridging: 
    \item[]\textbf{coreference}:  If an entity has a previous mention (and thus an IS of "Given"), it cannot be a bridging anaphor.
    \item[]\textbf{bridging contained}: The antecedent must be outside of the nominal phrase containing the anaphor, e.g., ``\textbf{the focus} of \underline{the story}'', should not be annotated as bridging. 
    \item[]\textbf{generic/world knowledge}: Entities that are accessible due to general world knowledge or situational context are not considered instances of bridging, e.g. ``the world'' or ``here'' should not be annotated as bridging.
    \item[]\textbf{explicit possessives}: If the anaphor contains an explicit possessive which corefers with the antecedent, no bridging relation is necessary, e.g., ``\underline{[Mark]}…\textbf{[his] house}'' should not be annotated as bridging.
\end{enumerate}

If an entity pair passes all 3 of the criteria above, then it is an instance of bridging, and the annotator proceeds to the subcategorization task. 

\subsection{Bridging Subtype Classification}

\begin{figure*}
    \centering
\resizebox{\textwidth}{!}{
\begin{tikzpicture}[
  level 1/.style={sibling distance=14em, level distance=6em},
  edge from parent/.style={->, thick, draw=black!70},
  edge from parent path={(\tikzparentnode.south) .. controls +(0,-0.7) and +(0,0.7) .. (\tikzchildnode.north)},
  >=latex,
  every node/.style={align=center, font=\sffamily}
]

% Root node
\node[draw, fill=blue!25, rounded corners, thick, text width=5cm, minimum height=1cm] (c0) {\textbf{\textsc{Bridging Subtypes}}}

% Level-1 children
  child {node[draw, fill=teal!20, rounded corners, text width=3cm, minimum height=0.8cm] (c1) {\textbf{\textsc{Comparison}}}}
  child {node[draw, fill=green!20, rounded corners, text width=3cm, minimum height=0.8cm] (c2) {\textbf{\textsc{Entity}}}}
  child {node[draw, fill=orange!25, rounded corners, text width=3cm, minimum height=0.8cm] (c3) {\textbf{\textsc{Set}}}}
  child {node[draw, fill=purple!20, rounded corners, text width=3cm, minimum height=0.8cm] (c4) {\textbf{\textsc{Other}}}};

% Sub-nodes
\begin{scope}[every node/.style={draw, fill=gray!15, rounded corners, font=\small\scshape, text width=2.5cm, minimum height=0.6cm}]
  % Comparison children
  \node [below of = c1, xshift=15pt] (c11) {relative};
  \node [below of = c11] (c12) {sense};
  \node [below of = c12] (c13) {time};

  % Entity children
  \node [below of = c2, xshift=15pt] (c21) {meronomy};
  \node [below of = c21] (c22) {property};
  \node [below of = c22] (c23) {resultative};
  \node [below of = c23] (c24) {associative};

  % Set children
  \node [below of = c3, xshift=15pt] (c31) {member};
  \node [below of = c31] (c32) {subset};
  \node [below of = c32] (c33) {span-interval};
\end{scope}

% Arrows
\foreach \value in {1,2,3}
  \draw[->, thick] (c1.195) |- (c1\value.west);

\foreach \value in {1,...,4}
  \draw[->, thick] (c2.195) |- (c2\value.west);
  
\foreach \value in {1,2,3}
  \draw[->, thick] (c3.195) |- (c3\value.west);

\end{tikzpicture}
}
    \caption{Bridging Subtype Classification in GUMBridge v1}
    \label{fig:subtype_taxonomy}
\end{figure*}
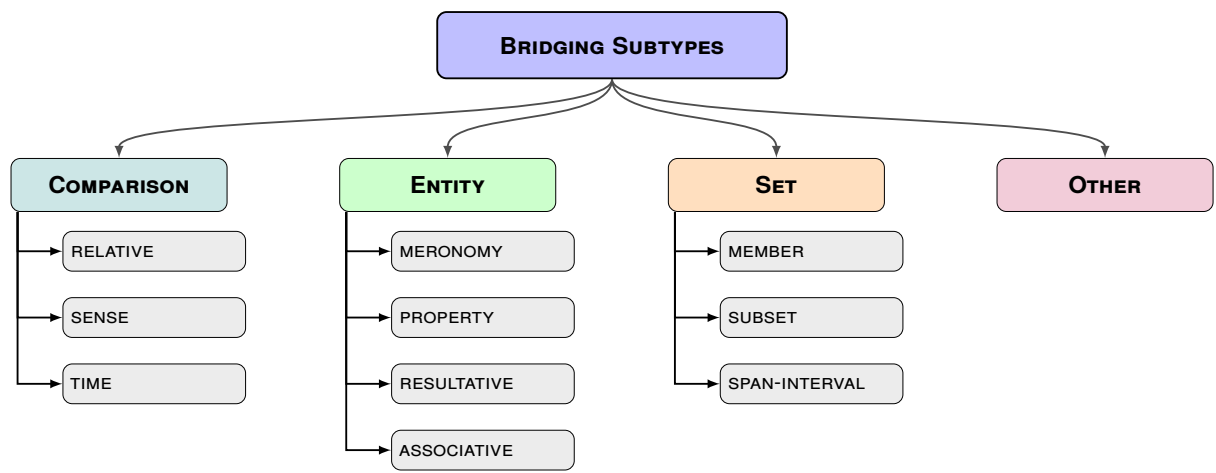

With the introduction of GUMBridge, we present a new subcategorization scheme for varieties of bridging anaphora. This system for subtype classification includes 3 main categories: \textsc{comparison} relations, \textsc{entity} relations, and \textsc{set} relations. Within these 3 categories, there are 10 subcategories, and there is an additional \textsc{other} category, for a total of 11 sub-varieties of bridging anaphora captured in GUMBridge. Figure \ref{fig:subtype_taxonomy} shows the full subcategorization scheme for GUMBridge v1. 

As the criteria for identifying instances of bridging is anaphoric, there is no theoretical need to limit the number of subtypes that can apply to an instance of bridging. The subtype labels are primarily a means of understanding how the phenomenon manifests in a discourse. As such, annotators were instructed to select all subtypes that apply to a bridging pair. To our knowledge, GUMBridge is the first English bridging resource to allow for such multi-subtype annotation. Table \ref{tab:subtype_counts} shows the counts of the bridging subtypes in the final version of GUMBridge v1. Below is a brief description and example of each bridging subtype (when multiple subtypes apply, it has been included in the footnotes): 

\begin{table}[]
\centering
\resizebox{0.7\columnwidth}{!}{%
\begin{tabular}{ll}
\toprule \toprule
\textbf{Subtypes} &  \\ \midrule
\hspace{2mm}\textbf{\textsc{Comparison}} &  \\
\hspace{5mm}\textsc{relative} & 1,200 \\
\hspace{5mm}\textsc{time} & 436 \\
\hspace{5mm}\textsc{sense} & 521 \\ \midrule
\hspace{5mm}Subtotal & 2,157 \\ \midrule
\hspace{2mm}\textbf{\textsc{Entity}} &  \\
\hspace{5mm}\textsc{associative} & 2,880 \\
\hspace{5mm}\textsc{meronomy} & 460 \\
\hspace{5mm}\textsc{property} & 148 \\
\hspace{5mm}\textsc{resultative} & 102 \\ \midrule
\hspace{5mm}Subtotal & 3,590 \\ \midrule
\hspace{2mm}\textbf{\textsc{Set}} &  \\
\hspace{5mm}\textsc{member} & 391 \\
\hspace{5mm}\textsc{subset} & 178 \\
\hspace{5mm}\textsc{span-interval} & 169 \\ \midrule
\hspace{5mm}Subtotal & 738 \\ \midrule
\hspace{2mm}\textbf{\textsc{Other}} & 67 \\ \midrule
\textbf{\textbf{Total Subtype Count}} & 6,552 \\ \midrule \midrule
\textbf{Multi-subtype Occurrences} &  \\ \midrule
\hspace{5mm}1 subtype & 4,887 \\
\hspace{5mm}2 subtypes & 770 \\
\hspace{5mm}3 subtypes & 39 \\ 
\hspace{5mm}4 subtypes & 2 \\ \midrule
\textbf{\textbf{Total Instances}} & 5,698 \\ \bottomrule \bottomrule 
\end{tabular}%
}
\caption{Bridging subtypes in GUMBridge v1.}
\label{tab:subtype_counts}
\end{table}

% A brief comparison to the bridging subtypes of ARRAU is included in Appendix X
% main relation types and their corresponding sub-categories

\paragraph{\textsc{comparison-relative}} The anaphor is preceded by a comparative marker (other, another, same, more, etc.), ordinal (second, third, etc.), comparative adjective (larger, smaller, etc.), or superlative (best, worst, etc.) which implies a comparison to the antecedent (or vice versa). 

\ex. \underline{Several women} walked into the room. \textbf{Other women} soon followed.

\paragraph{\textsc{comparison-sense}} The type of the anaphor is omitted but inferable via comparison to the antecedent (or vice versa).

\ex. I’ve been to the \underline{Chinese restaurant}. I want to go to \textbf{the Italian one}.

\paragraph{\textsc{comparison-time}} The anaphor refers to a specific time/time frame which is understandable with reference to the time/time frame expressed by the antecedent (or vice versa). 

\ex. I went shopping \underline{Wednesday, March 3rd}. I will go again \textbf{the following Wednesday}.\footnote{also \textsc{comparison-relative}}

\paragraph{\textsc{entity-meronomy}} The anaphor is a subunit of the antecedent (or vice versa), i.e., there is some part-whole relation between the anaphor and the antecedent, including physical subparts, substance-portion, and regions/subsections.

\ex. I saw \underline{a large house} by the lake. \textbf{The door} was red. 

\paragraph{\textsc{entity-property}} The anaphor is a physical or intangible property of the antecedent (or vice versa). For example: smell, length, style, etc.

\ex. I picked up \underline{a bouquet of roses}. \textbf{The scent} was lovely. 

\paragraph{\textsc{entity-resultative}} The anaphor is logically inferable from the antecedent (or vice versa). This is often the result of a transformative/product producing process, like cooking/baking.\footnote{This covers the \textsc{transformed} type proposed by \citet{fang-etal-2022-take} specifically for recipe outcomes.}

\ex. Though \underline{my flour} was a strange texture, \textbf{the bread} came out perfectly.

%\ex. Though \underline{my dough} was a strange texture, \textbf{the bread} came out perfectly.

\paragraph{\textsc{entity-associative}} The anaphor is an attribute or closely associated entity of the antecedent (or vice versa). This frequently manifests as implicit arguments of a predicate \ref{ex:arg}, relational nouns \ref{ex:relational}, and prototypical associations \ref{ex:prototype}: 

\ex. There was \underline{a murder} last night. \textbf{The victim} has yet to be identified.
\label{ex:arg}

\ex. There is \underline{a child} in the park. \textbf{The parent} must be nearby.
\label{ex:relational}

\ex. I went to \underline{a wedding} last week. \textbf{The reception} was really fun.
\label{ex:prototype}

\paragraph{\textsc{set-member}} The anaphor is an element of the antecedent set (or vice versa). This includes group-member and class-instance relations.

\ex. I got \underline{several books} for my birthday. \textbf{The mystery novel} was my favorite. 

\paragraph{\textsc{set-subset}} The anaphor is a subset of the antecedent set (or vice versa).

\ex. \underline{A group of students} entered the hall. \textbf{The boys} wore neckties with their uniforms. 

\paragraph{\textsc{set-span-interval}} The anaphor is a sub-span of the spatial or temporal antecedent interval (or vice versa). 

\ex. If you want to meet up on \underline{Sunday}, I will be free in \textbf{the morning}.\footnote{also \textsc{comparison-time}}

\paragraph{\textsc{other}} The  \textsc{other} category is for instances which fit the information status based definition of a bridging pair but do not fall into any of the bridging subtype categories outlined above. For example, \citet{ogrodniczuk-zawislawska-2016-bridging} discuss metareference, which allows for reference back to a name or label:
%and temporal-dissimilation:

\ex. I went to \underline{Sensational Cakes} yesterday, but I didn't think \textbf{the cakes} were very good.
\label{ex:meta}

While such instances are not numerous enough to warrant additional subtype categories, they are still unique and worth capturing for future investigation. When we consider the distribution of subtypes shown in Table \ref{tab:subtype_counts}, we see that only 67 of the 5,698 instances of bridging in GUMBridge are labeled with \textsc{other}. This indicates that the subtype scheme provides good coverage for the varieties of bridging anaphora present in the data. 

However, it is notable that some subtypes are far more frequent than others. For example, 51\% (2,880 instances) of the bridging pairs are labeled with the \textsc{entity-associative} subtype, while only 9\% (521 instances) are labeled with \textsc{comparison-sense}. While differences in the distribution of subtype categories are linguistically interesting and worth investigating, it is also worth noting that \textsc{entity-associative} is the most broadly construed subtype, and can manifest in a variety of linguistic ways (prototypical associations, implicit possessives, implicit arguments, etc.). 
While we choose to group certain varieties together into one subtype category because they are difficult to reliably distinguish, we acknowledge that in the future it may be worthwhile to further delineate larger subtype categories, like \textsc{entity-associative}.

%\ex. \underline{Several women} walked into the room. \textbf{One} left immediately.
%\label{ex:sense_member}

%\ex. I will come to visit \underline{this week}, as I could not come \textbf{the previous week}.
%\label{ex:relative_time}

%Example \ref{ex:sense_member} shows an instance for which \textsc{comparison-sense} and \textsc{set-member} both apply, while example \ref{ex:relative_time} show a case where \textsc{comparison-relative} and \textsc{comparison-time} apply. In this annotation pilot, annotators where instructed to select a single bridging subtype, prioritizing certain categories over others if they occurred together. However, in principle, all applicable subtypes could be annotated. In our subsequent efforts to annotate the remaining data in GUM and produce a full version of GUMBridge, we intend to support the annotation of multiple bridging subtypes for a single bridging pair for the entire corpus. 

\section{Inter-Annotator Agreement Study}
\label{sec:IAA}

% test and dev, agreement numbers, cite subjectivity paper

\begin{table*}[]
\centering
\resizebox{0.9\textwidth}{!}{%
\begin{tabular}{clccclclcc}
\toprule
\multirow{2}{*}{\textbf{Annotator Pairs}} &  & \multicolumn{3}{c}{\textbf{\begin{tabular}[c]{@{}c@{}}Anaphor \\ Recognition\end{tabular}}} &  & \textbf{\begin{tabular}[c]{@{}c@{}}Antecedent \\ Resolution\end{tabular}} &  & \multicolumn{2}{c}{\textbf{\begin{tabular}[c]{@{}c@{}}Bridging \\ Subtype\end{tabular}}} \\ \cmidrule{3-5} \cmidrule{7-7} \cmidrule{9-10} 
 &  & \textbf{Precision} & \textbf{Recall} & \textbf{F1 Score} &  & \textbf{Accuracy} &  & \textbf{Accuracy} & \textbf{Cohen's $\kappa$} \\ %\cline{1-1} \cline{3-5} \cline{7-7} \cline{9-10} 
 \midrule
A &  & 0.84 & \textbf{0.81} & \textbf{0.83} &  & \textbf{0.94} &  & \textbf{0.90} & \textbf{0.86} \\
B &  & \textbf{0.93} & 0.68 & 0.79 &  & 0.90 &  & 0.82 & 0.80 \\
C &  & 0.66 & 0.36 & 0.47 &  & 0.70 &  & 0.50 & 0.58 \\
D &  & 0.22 & 0.27 & 0.24 &  & 0.45 &  & 0.60 & 0.58 \\ \cmidrule{1-1} \cmidrule{3-5} \cmidrule{7-7} \cmidrule{9-10} 
Micro Avg. &  & 0.69 & 0.56 & 0.62 &  & 0.84 &  & 0.79 & 0.76 \\ \bottomrule
\end{tabular}%
}
\caption{GUMBridge dev pilot inter-annotator agreement.}
\label{tab:agree}
\end{table*}

In order to validate the robustness of our annotation procedure and proposed schema for subtype categorization, we conducted an 
Inter-Annotator Agreement study on the dev set of GUMBridge. Each of the 32 documents in the dev set (\textasciitilde30k tokens) was double annotated by 4 different pairs of annotators, with each pair of annotators completing 8 documents. We note that the annotators in pair B are the authors of this paper, and therefore the most expert annotators in the pool. The other pairs of annotators are linguistics graduate students who completed several hours of training prior to the pilot. The results of this study are shown in Table~\ref{tab:agree}. We provide agreement numbers for each stage of annotation in our procedure: (1) recognition of bridging anaphora, (2) resolution back to the associative antecedent, and (3) subcategorization of the bridging pair. We provide the aggregate scores for the annotators together, as well as the scores of the individual pairs. 

For the recognition of the bridging anaphor, we give mutual P/R/F of Annotator 1 relative to Annotator 2. The micro avg. F1 for bridging anaphor recognition is only 0.62, indicating that anaphor recognition is a difficult subtask. As the selection of a correct antecedent is limited by the recognition of the anaphor, we give the accuracy of Annotator 2 selecting the antecedent entity when both annotators agree on the bridging anaphor, which is 84\% of a total of 336 cases. For the bridging subtype scores, accuracy shows the proportion of exact matches, while Cohen's $\kappa$ is reported on the individual subtype judgments (e.g., a bridging pair with 2 subtype annotations is counted as 2 annotator judgments). For the 283 instances where both annotators agreed on the anaphor and antecedent of a bridging pair, the Cohen's $\kappa$ for the bridging subtype annotation is 0.76, which indicates substantial agreement. From these scores, we see that the detection of bridging anaphora is the limiting factor in producing reliable annotations for instances of bridging.

Looking at the individual annotator pair scores, we see very clear divides in performance. While pairs A and B achieve high agreement on all tasks, pairs C and D perform substantially worse, particularly on the task of anaphor recognition. This inconsistency is in line with recent work which has shown annotation of bridging anaphora to be a highly subjective task, dependent on an annotator's personal judgment of how entities are related in a discourse \citep{levine-zeldes-2025-subjectivity}. We note that our most expert annotators (pair B) were able to achieve good agreement. Amongst the 3 other pairs, who all received the same amount of training, the agreement ranged from high (pair A), to moderate (pair C), to low (pair D). From these results, we see that the various subtasks of bridging annotation are possible to learn and agree on, particularly for expert annotators. However, it is still a very difficult set of tasks, and it cannot be taken for granted that all annotators will perform reliably.

\begin{figure}
  \centering
  \includegraphics[width=1\linewidth]{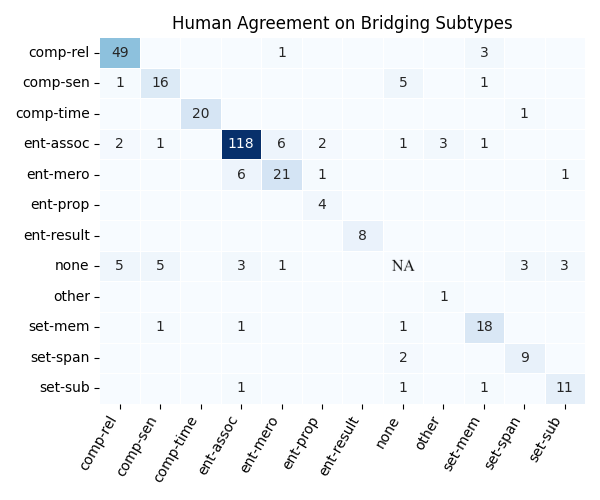}
  \caption{Confusion matrix of bridging subtypes for bridging instances with matching anaphor and antecedent annotations.}
  \label{fig:bridgetype_cm_human}
\end{figure}

In Figure~\ref{fig:bridgetype_cm_human}, we show a confusion matrix of the bridging subtype labels assigned by annotators to overlapping bridging pairs (bridging instances agreed on by both annotators in an annotator pair). We see that while there is overlap for a number of different subtypes, most confusions only occur one or twice. The subtypes with the greatest overlap are \textsc{entity-meronomy} and \textsc{entity-associative}, with a total of 12 disagreements between the two categories. With a Cohen's $\kappa$ of 0.76 for the aggregate of annotator performance, we see the schema for bridging type classification presented here is robust, but it is clear from the scattering of disagreements that some clarification on confusable edge cases would be beneficial. 

%However, it is worth noting, and also it is clear that entity associative is an overpopulated category, which should be further divided in the future. 

%\subsection{Corpus Statistics}

% stats table - # documents, tokens, mentions, 
% subtype distribution (by genre?)

\section{LLM Evaluation}\label{sec:LLM}

\begin{table*}[t!]
\centering
\resizebox{0.9\textwidth}{!}{%
\begin{tabular}{clccclclcc}
\toprule
\multirow{2}{*}{\textbf{Model}} &  & \multicolumn{3}{c}{\textbf{\begin{tabular}[c]{@{}c@{}}Anaphor \\ Recognition\end{tabular}}} &  & \textbf{\begin{tabular}[c]{@{}c@{}}Antecedent \\ Resolution\end{tabular}} &  & \multicolumn{2}{c}{\textbf{\begin{tabular}[c]{@{}c@{}}Bridging \\ Subtype\end{tabular}}} \\ \cmidrule{3-5} \cmidrule{7-7} \cmidrule{9-10} 
 &  & \textbf{Precision} & \textbf{Recall} & \textbf{F1 Score} &  & \textbf{Accuracy} &  & \textbf{Accuracy} & \textbf{Cohen's $\kappa$} \\ 
 \midrule
 %\cline{1-1} \cline{3-5} \cline{7-7} \cline{9-10} 
GPT-5 & \multicolumn{1}{c}{} & \textbf{0.43} & \textbf{0.38} & \textbf{0.40} & \multicolumn{1}{c}{} & \textbf{0.49} & \multicolumn{1}{c}{} & \textbf{0.69} & \textbf{0.64} \\ 
Llama-3.3 & \multicolumn{1}{c}{} & 0.15 & 0.30 & 0.20 & \multicolumn{1}{c}{} & 0.28 & \multicolumn{1}{c}{} & 0.61 & 0.27 \\ 
Qwen3 & \multicolumn{1}{c}{} & 0.14 & 0.21 & 0.17 & \multicolumn{1}{c}{} & 0.17 & \multicolumn{1}{c}{} & 0.53 & 0.34 \\ 
\bottomrule
\end{tabular}%
}
\caption{Results of LLM baselines on GUMBridge test.}
\label{tab:baseline}
\end{table*}

% set up of LLM zero and few shot prompting evalutation
% Performance results

%\section{Error Analysis}

% Comparison of LLM results to human results -> which subtypes where easiest for humans/LLMs to identify and resolve back to -> conf matrix, genre differnces?, 

In this section, we provide an LLM baseline for the task of bridging resolution evaluated on the test set of GUMBridge (32 documents, \textasciitilde30k tokens, 562 instances of bridging). We provide scores for two open models \texttt{meta-llama/llama-3.3-70b-instruct} and \texttt{qwen/qwen3-235b-a22b-instruct-2507}, and one closed SoTA model, \texttt{GPT-5}. Following the set up of the evaluation of our human annotators, we test the models on 3 subtasks: (1) anaphora recognition, (2) antecedent selection, and (3) subtype categorization. For each subtask, the models are provided with a separate prompt which gives a detailed explanation of the task with a series of few-shot examples.\footnote{Templates of the exact prompts are included in Appendix \ref{sec:appendix_prompts}} For the anaphor recognition subtask, the models are queried sentence by sentence through the document, with 50 tokens of additional context on each side of the sentence. For the antecedent selection subtask, the models are queried once for each bridging anaphor in the gold annotations, getting the sentence containing the anaphor with 150 tokens of preceding context. For the subtype categorization subtask, the models are queried once for each bridging pair in the gold annotations, getting the sentences containing the anaphor and the antecedent, each with 50 tokens of additional context on each side. The results of this evaluation are shown in Table~\ref{tab:baseline}. The reported scores are from single runs on each model. 

\begin{figure}
  \centering
  \includegraphics[width=1\linewidth]{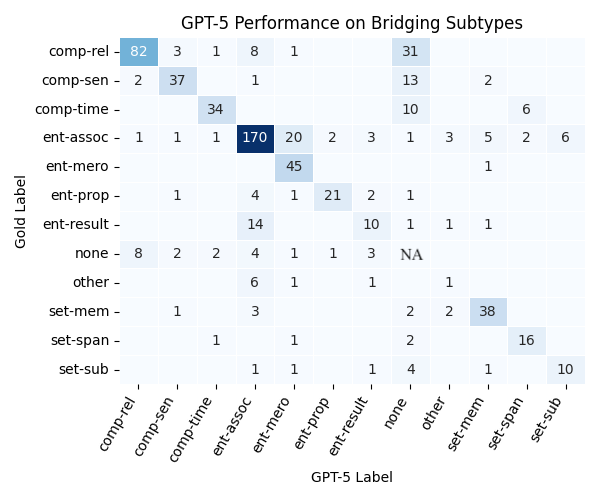}
  \caption{Confusion matrix of bridging subtypes for \texttt{GPT-5} predicted labels.}
  \label{fig:bridgetype_cm_llm}
\end{figure}

%\begin{table} %[h!tb]
%\centering
%\resizebox{\columnwidth}{!}{%
%\begin{tabular}{ccccc}
%\hline
% & \textbf{Model} & \textbf{Precision} & \textbf{Recall} & \textbf{F1 Score} \\ \hline
%\multirow{2}{*}{\begin{tabular}[c]{@{}c@{}}Anaphor\\Recognition\end{tabular}} 
% & GPT-5 & 0.45 & 0.38 & 0.41 \\
% & Llama & 0.15 & 0.31 & 0.21 \\ \hline
% & & \multicolumn{2}{c}{\textbf{Accuracy}} & \\ \hline
%\multirow{2}{*}{\begin{tabular}[c]{@{}c@{}}Antecedent\\Resolution\end{tabular}}% 
% & GPT-5 & \multicolumn{2}{c}{0.50} & \\
% & Llama & \multicolumn{2}{c}{0.28} & \\ \hline
% & & \multicolumn{2}{c}{\textbf{Accuracy}} & \textbf{Cohen's $\kappa$} \\ \hline
%\multirow{2}{*}{\begin{tabular}[c]{@{}c@{}}Bridging\\Subtype\end{tabular}} 
% & GPT-5 & \multicolumn{2}{c}{0.69} & 0.64 \\
% & Llama & \multicolumn{2}{c}{0.61} & 0.27 \\ \hline
%\end{tabular}%
%}
%\caption{Results of LLM baselines (GPT-5 and Open) on GUMBridge test.}
%\label{tab:llm-baseline}
%\end{table}

While \texttt{GPT-5} performs substantially worse than the most expertly trained annotators, we see that it scores substantially better than under-performing annotators. Comparing the results of \texttt{GPT-5} to the aggregate of the human annotators, we see that the scores are lower overall. However, considering that the SoTA scores on bridging resolution (for anaphor recognition and antecedent selection together) are under 0.5 P/R/F for evalution on ISNotes, BASHI, and ARRAU \cite{kobayashi-etal-2023-pairspanbert, kobayashi-ng-2020-bridging}, we consider this \texttt{GPT-5} baseline to be reasonably impressive. On the other hand, \texttt{llama-3.3-70b-instruct} and \texttt{qwen3-235b-a22b-instruct-2507} perform quite poorly at the tasks of anaphor recognition (F1 0.20 and 0.17 respectively) and antecedent selection (Acc 0.28 and 0.17 respectively), and perform adequately at subtype categorization in terms of accuracy (0.61 and 0.53 respectively). This suggests that models of this size/capacity may not have the ability to recognize bridging pairs in context. While bridging resolution and subtype categorization remain difficult NLP tasks, even while leveraging LLMs, the results of the \texttt{GPT-5} baseline suggest that LLMs have the capacity be a useful tool in improving the performance of bridging resolution systems.

In Figure~\ref{fig:bridgetype_cm_llm} we provide a confusion matrix of the bridging subtype labels assigned by \texttt{GPT-5} and the gold labels from the test set of GUMBridge. Compared with the results of the human annotators in Figure~\ref{fig:bridgetype_cm_human}, we see that there are considerably more categories being confused, but most of the disagreements still only occur a handful of times. The subtypes which were most commonly confused, with 20 instances, are \textsc{entity-meronomy} and \textsc{entity-associative}, which was also the case for the human annotation. Additionally, we see that \texttt{GPT-5} misses a number of instances for \textsc{comparison} relations in cases when they co-occur with other subtype labels. 

\begin{figure}[h!tb]
  \centering
  \includegraphics[width=.9\linewidth]{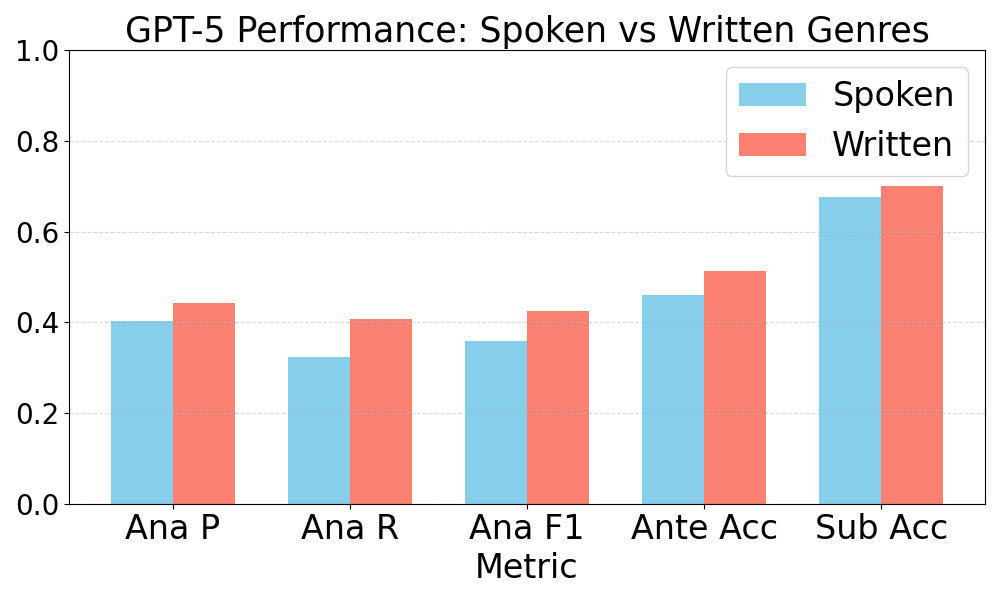}
\vspace{-10pt}
\caption{Performance of \texttt{GPT-5} on spoken genres vs. written genres.}
  \label{fig:genre_comparison}
\end{figure}

% (court, conversation, interview, podcast, speech, vlog)
% (academic, bio, essay, fiction, letter, news, reddit, textbook, voyage, whow)

In Figure~\ref{fig:genre_comparison} we compare the performance of \texttt{GPT-5} on spoken vs. written genres for our evaluations metrics. To our knowledge, this is the first cross-genre comparison of performance on the task of bridging resolution for English (while the sub-corpora of ARRAU cover multiple genres, evaluation has primarily been conducted on ARRAU RST's WSJ news data alone). Looking across the different metrics, we consistently see that spoken genres are more difficult for the model than written genres. We also see that the relative difficulty of the subtasks remains consistent between written and spoken data (anaphor recognition is the hardest, followed by antecedent selection, and subtype categorization is the easiest). While brief, we include this analysis to illustrate the important types of linguistic research which will now be possible with the additional genre diversity provided by GUMBridge.

\section{Conclusion}

% Resource summary, value of this resource for future work - linguistic analysis for bridging anaphora, evaluation of natural language understanding/discourse level understanding in LLMs, harmonization of existing bridging resources/standardization of bridging resolution task

In this paper, we introduce GUMBridge, a new English resource for varieties of bridging anaphora. Taking an information status informed perspective, we provide broad coverage, having one of the highest densities of bridging annotation in available English resources (19.6 instances per 1k tokens). We also provide subtype annotations for sub-varieties of bridging anaphora, which will allow for fine-grained linguistic analysis of how bridging instances manifest in a discourse. We note that this is the first English bridging resource to formally recognize the theoretical compatibility of multiple subtypes within a single instance and to allow for multi-subtype annotation. This dataset is the largest English resource available for instances of bridging (containing more bridging instances than the combined sub-corpora of ARRAU), particularly for referential bridging (containing more than 4 times the bridging instances of ISNotes and BASHI combined). GUMBridge also has by far the greatest range of genre diversity, covering 24 different genres, including both written and spoken data. 

We also provide an LLM baseline for bridging resolution and subtype categorization evaluated on the test set of GUMBridge. Results from this evaluation indicate that bridging related tasks remain difficult even with the use of SoTA LLMs, achieving only an F1 Score of 0.40 for the task of anaphor recognition. However, given the low performance of current SoTA bridging resolution systems, the baseline also indicates the potential of LLMs to help improve bridging resolution systems in the future.

\section*{Limitations}

This work is limited by the fact that GUMBridge contains only English language data, and more work will need to be done to address bridging in a multilingual capacity. While the information status based approach to identify bridging anaphora is largely language agnostic, the subtype categories of GUMBridge were developed with English data in mind and may not straightforwardly extend to other languages. Additionally, the LLM baselines included in this paper are minimal, only testing three models in a single configuration, reporting scores from a single run. This baseline is meant to illustrate that bridging resolution is still an open, challenging task, rather than to present a full system for bridging resolution.

%\nocite{Eco:1990,Martin-90,Chercheur-94,CastorPollux-92,bs-2570-manual,bs-2570-techreport,chomsky-73,hoel-71-portion,hoel-71-whole,singer-whole,Jespersen:1922,Superman-Batman-Catwoman-Spiderman-00}
\section*{Bibliographical References}\label{sec:reference}

\bibliographystyle{lrec2026-natbib}
\bibliography{lrec2026-example}

\begin{thebibliography}{11}
\expandafter\ifx\csname natexlab\endcsname\relax\def\natexlab#1{#1}\fi

\bibitem[{Aoyama et~al.(2023)Aoyama, Behzad, Gessler, Levine, Lin, Liu, Peng, Zhu, and Zeldes}]{aoyama-etal-2023-gentle}
Aoyama, Tatsuya and Behzad, Shabnam and Gessler, Luke and Levine, Lauren and Lin, Jessica and Liu, Yang Janet and Peng, Siyao and Zhu, Yilun and Zeldes, Amir. 2023.
\newblock \href {https://doi.org/10.18653/v1/2023.law-1.17} {\emph{{GENTLE}: A Genre-Diverse Multilayer Challenge Set for {E}nglish {NLP} and Linguistic Evaluation}}.
\newblock Association for Computational Linguistics.

\bibitem[{Eckart et~al.(2012)Eckart, Riester, and Schweitzer}]{Eckart2012}
Eckart, Kerstin and Riester, Arndt and Schweitzer, Katrin. 2012.
\newblock \href {https://doi.org/10.1007/978-3-642-28249-2_7} {\emph{A Discourse Information Radio News Database for Linguistic Analysis}}.
\newblock Springer Berlin Heidelberg.

\bibitem[{Markert et~al.(2012)Markert, Hou, and Strube}]{markert-etal-2012-collective}
Markert, Katja and Hou, Yufang and Strube, Michael. 2012.
\newblock \href {https://aclanthology.org/P12-1084/} {\emph{Collective Classification for Fine-grained Information Status}}.
\newblock Association for Computational Linguistics.

\bibitem[{Nedoluzhko et~al.(2009)Nedoluzhko, M{\'i}rovsk{\'y}, and Pajas}]{nedoluzhko-etal-2009-coding}
Nedoluzhko, Anna and M{\'i}rovsk{\'y}, Ji{\v{r}}{\'i} and Pajas, Petr. 2009.
\newblock \href {https://aclanthology.org/W09-3017/} {\emph{The Coding Scheme for Annotating Extended Nominal Coreference and Bridging Anaphora in the {P}rague Dependency Treebank}}.
\newblock Association for Computational Linguistics, ISLRN \href{https://www.islrn.org/resources/942-053-729-014-3}{942-053-729-014-3}.

\bibitem[{Ogrodniczuk and Zawis{\l}awska(2016)}]{ogrodniczuk-zawislawska-2016-bridging}
Ogrodniczuk, Maciej and Zawis{\l}awska, Magdalena. 2016.
\newblock \href {https://doi.org/10.18653/v1/W16-0703} {\emph{Bridging Relations in {P}olish: Adaptation of Existing Typologies}}.
\newblock Association for Computational Linguistics.

\bibitem[{Poesio and Artstein(2008)}]{poesio-artstein-2008-anaphoric}
Poesio, Massimo and Artstein, Ron. 2008.
\newblock \href {https://aclanthology.org/L08-1091/} {\emph{Anaphoric Annotation in the {ARRAU} Corpus}}.
\newblock European Language Resources Association (ELRA), ISLRN \href{https://www.islrn.org/resources/462-157-606-044-8}{462-157-606-044-8}.

\bibitem[{R{\"o}siger(2018)}]{rosiger-2018-bashi}
R{\"o}siger, Ina. 2018.
\newblock \href {https://aclanthology.org/L18-1058/} {\emph{{BASHI}: A Corpus of {W}all {S}treet {J}ournal Articles Annotated with Bridging Links}}.
\newblock European Language Resources Association (ELRA).

\bibitem[{Schweitzer et~al.(2018)Schweitzer, Eckart, G{\"a}rtner, Falenska, Riester, R{\"o}siger, Schweitzer, Stehwien, and Kuhn}]{schweitzer-etal-2018-german}
Schweitzer, Katrin and Eckart, Kerstin and G{\"a}rtner, Markus and Falenska, Agnieszka and Riester, Arndt and R{\"o}siger, Ina and Schweitzer, Antje and Stehwien, Sabrina and Kuhn, Jonas. 2018.
\newblock \href {https://aclanthology.org/L18-1457/} {\emph{{G}erman Radio Interviews: The {GRAIN} Release of the {SFB}732 Silver Standard Collection}}.
\newblock European Language Resources Association (ELRA).

\bibitem[{Uryupina et~al.(2019)Uryupina, Artstein, Bristot, Cavicchio, Delogu, Rodr{\'i}guez, and Poesio}]{Uryupina2019AnnotatingAB}
Olga Uryupina and Ron Artstein and Antonella Bristot and Federica Cavicchio and Francesca Delogu and Kepa Joseba Rodr{\'i}guez and Massimo Poesio. 2019.
\newblock \href {https://api.semanticscholar.org/CorpusID:164858637} {\emph{Annotating a broad range of anaphoric phenomena, in a variety of genres: the ARRAU Corpus}}.

\bibitem[{Weischedel et~al.(2012)Weischedel, Pradhan, Ramshaw, Kaufman, Franchini, El-Bachouti, Xue, Palmer, Hwang, Bonial, Choi, Mansouri, Foster, aati Hawwary, Marcus, Taylor, Greenberg, Hovy, Belvin, and Houston}]{WeischedelPradhanRamshawEtAl2012}
Ralph Weischedel and Sameer Pradhan and Lance Ramshaw and Jeff Kaufman and Michelle Franchini and Mohammed El-Bachouti and Nianwen Xue and Martha Palmer and Jena D. Hwang and Claire Bonial and Jinho Choi and Aous Mansouri and Maha Foster and Abdel-aati Hawwary and Mitchell Marcus and Ann Taylor and Craig Greenberg and Eduard Hovy and Robert Belvin and Ann Houston. 2012.
\newblock \emph{OntoNotes Release 5.0}.

\bibitem[{Zeldes(2017)}]{Zeldes2017}
Amir Zeldes. 2017.
\newblock \href {https://doi.org/http://dx.doi.org/10.1007/s10579-016-9343-x} {\emph{The {GUM} Corpus: Creating Multilayer Resources in the Classroom}}.
\newblock ISLRN \href{https://www.islrn.org/resources/421-566-418-865-2}{421-566-418-865-2}.

\end{thebibliography}


\begin{thebibliography}{22}
\expandafter\ifx\csname natexlab\endcsname\relax\def\natexlab#1{#1}\fi

\bibitem[{Asher and Lascarides(1998)}]{asher1998bridging}
Nicholas Asher and Alex Lascarides. 1998.
\newblock Bridging.
\newblock \emph{Journal of Semantics}, 15(1):83--113.

\bibitem[{Baumann and Riester(2012)}]{baumann2012referential}
Stefan Baumann and Arndt Riester. 2012.
\newblock Referential and lexical givenness: Semantic, prosodic and cognitive aspects.
\newblock \emph{Prosody and meaning}, 25:119--162.

\bibitem[{Clark(1975)}]{clark-1975-bridging}
Herbert~H. Clark. 1975.
\newblock \href {https://aclanthology.org/T75-2034/} {Bridging}.
\newblock In \emph{Theoretical Issues in Natural Language Processing}.

\bibitem[{Fang et~al.(2022)Fang, Baldwin, and Verspoor}]{fang-etal-2022-take}
Biaoyan Fang, Timothy Baldwin, and Karin Verspoor. 2022.
\newblock \href {https://doi.org/10.18653/v1/2022.findings-acl.275} {What does it take to bake a cake? the {R}ecipe{R}ef corpus and anaphora resolution in procedural text}.
\newblock In \emph{Findings of the Association for Computational Linguistics: ACL 2022}, pages 3481--3495, Dublin, Ireland. Association for Computational Linguistics.

\bibitem[{Hawkins(1978)}]{hawkins1978definiteness}
John~A. Hawkins. 1978.
\newblock Definiteness and indefiniteness: A study in reference and grammaticality prediction.
\newblock \emph{Journal of Linguistics}, 27:405–442.

\bibitem[{Hou(2020)}]{hou-2020-bridging}
Yufang Hou. 2020.
\newblock \href {https://doi.org/10.18653/v1/2020.acl-main.132} {Bridging anaphora resolution as question answering}.
\newblock In \emph{Proceedings of the 58th Annual Meeting of the Association for Computational Linguistics}, pages 1428--1438, Online. Association for Computational Linguistics.

\bibitem[{Khosla et~al.(2021)Khosla, Yu, Manuvinakurike, Ng, Poesio, Strube, and Ros{\'e}}]{khosla-etal-2021-codi}
Sopan Khosla, Juntao Yu, Ramesh Manuvinakurike, Vincent Ng, Massimo Poesio, Michael Strube, and Carolyn Ros{\'e}. 2021.
\newblock \href {https://doi.org/10.18653/v1/2021.codi-sharedtask.1} {The {CODI}-{CRAC} 2021 shared task on anaphora, bridging, and discourse deixis in dialogue}.
\newblock In \emph{Proceedings of the CODI-CRAC 2021 Shared Task on Anaphora, Bridging, and Discourse Deixis in Dialogue}, pages 1--15, Punta Cana, Dominican Republic. Association for Computational Linguistics.

\bibitem[{Kobayashi et~al.(2023)Kobayashi, Hou, and Ng}]{kobayashi-etal-2023-pairspanbert}
Hideo Kobayashi, Yufang Hou, and Vincent Ng. 2023.
\newblock \href {https://doi.org/10.18653/v1/2023.acl-long.383} {{P}air{S}pan{BERT}: An enhanced language model for bridging resolution}.
\newblock In \emph{Proceedings of the 61st Annual Meeting of the Association for Computational Linguistics (Volume 1: Long Papers)}, pages 6931--6946, Toronto, Canada. Association for Computational Linguistics.

\bibitem[{Kobayashi and Ng(2020)}]{kobayashi-ng-2020-bridging}
Hideo Kobayashi and Vincent Ng. 2020.
\newblock \href {https://doi.org/10.18653/v1/2020.coling-main.331} {Bridging resolution: A survey of the state of the art}.
\newblock In \emph{Proceedings of the 28th International Conference on Computational Linguistics}, pages 3708--3721, Barcelona, Spain (Online). International Committee on Computational Linguistics.

\bibitem[{Levine and Zeldes(2024)}]{levine-zeldes-2024-unifying}
Lauren Levine and Amir Zeldes. 2024.
\newblock \href {https://doi.org/10.18653/v1/2024.crac-1.5} {Unifying the scope of bridging anaphora types in {E}nglish: Bridging annotations in {ARRAU} and {GUM}}.
\newblock In \emph{Proceedings of the Seventh Workshop on Computational Models of Reference, Anaphora and Coreference}, pages 41--51, Miami. Association for Computational Linguistics.

\bibitem[{Levine and Zeldes(2025)}]{levine-zeldes-2025-subjectivity}
Lauren Levine and Amir Zeldes. 2025.
\newblock \href {https://doi.org/10.18653/v1/2025.law-1.4} {Subjectivity in the annotation of bridging anaphora}.
\newblock In \emph{Proceedings of the 19th Linguistic Annotation Workshop (LAW-XIX-2025)}, pages 48--59, Vienna, Austria. Association for Computational Linguistics.

\bibitem[{Lin and Zeldes(2021)}]{lin-zeldes-2021-wikigum}
Jessica Lin and Amir Zeldes. 2021.
\newblock \href {https://doi.org/10.18653/v1/2021.law-1.18} {{W}iki{GUM}: Exhaustive entity linking for wikification in 12 genres}.
\newblock In \emph{Proceedings of the Joint 15th Linguistic Annotation Workshop (LAW) and 3rd Designing Meaning Representations (DMR) Workshop}, pages 170--175, Punta Cana, Dominican Republic. Association for Computational Linguistics.

\bibitem[{Nissim et~al.(2004)Nissim, Dingare, Carletta, and Steedman}]{nissim-etal-2004-annotation}
Malvina Nissim, Shipra Dingare, Jean Carletta, and Mark Steedman. 2004.
\newblock \href {https://aclanthology.org/L04-1402/} {An annotation scheme for information status in dialogue}.
\newblock In \emph{Proceedings of the Fourth International Conference on Language Resources and Evaluation ({LREC}{'}04)}, Lisbon, Portugal. European Language Resources Association (ELRA).

\bibitem[{Ogrodniczuk and Zawis{\l}awska(2016)}]{ogrodniczuk-zawislawska-2016-bridging}
Maciej Ogrodniczuk and Magdalena Zawis{\l}awska. 2016.
\newblock \href {https://doi.org/10.18653/v1/W16-0703} {Bridging relations in {P}olish: Adaptation of existing typologies}.
\newblock In \emph{Proceedings of the Workshop on Coreference Resolution Beyond {O}nto{N}otes ({CORBON} 2016)}, pages 16--22, San Diego, California. Association for Computational Linguistics.

\bibitem[{Prince(1981)}]{prince1981toward}
Ellen~F. Prince. 1981.
\newblock Toward a taxonomy of given-new information.
\newblock \emph{Radical pragmatics}, pages 223--255.

\bibitem[{Roesiger et~al.(2018)Roesiger, Riester, and Kuhn}]{roesiger-etal-2018-bridging}
Ina Roesiger, Arndt Riester, and Jonas Kuhn. 2018.
\newblock \href {https://aclanthology.org/C18-1298/} {Bridging resolution: Task definition, corpus resources and rule-based experiments}.
\newblock In \emph{Proceedings of the 27th International Conference on Computational Linguistics}, pages 3516--3528, Santa Fe, New Mexico, USA. Association for Computational Linguistics.

\bibitem[{R{\"o}siger(2018)}]{rosiger-2018-bashi}
Ina R{\"o}siger. 2018.
\newblock \href {https://aclanthology.org/L18-1058/} {{BASHI}: A corpus of {W}all {S}treet {J}ournal articles annotated with bridging links}.
\newblock In \emph{Proceedings of the Eleventh International Conference on Language Resources and Evaluation ({LREC} 2018)}, Miyazaki, Japan. European Language Resources Association (ELRA).

\bibitem[{Yu et~al.(2022)Yu, Khosla, Manuvinakurike, Levin, Ng, Poesio, Strube, and Ros{\'e}}]{yu-etal-2022-codi}
Juntao Yu, Sopan Khosla, Ramesh Manuvinakurike, Lori Levin, Vincent Ng, Massimo Poesio, Michael Strube, and Carolyn Ros{\'e}. 2022.
\newblock \href {https://aclanthology.org/2022.codi-crac.1/} {The {CODI}-{CRAC} 2022 shared task on anaphora, bridging, and discourse deixis in dialogue}.
\newblock In \emph{Proceedings of the CODI-CRAC 2022 Shared Task on Anaphora, Bridging, and Discourse Deixis in Dialogue}, pages 1--14, Gyeongju, Republic of Korea. Association for Computational Linguistics.

\bibitem[{Zeldes(2022)}]{Zeldes2022}
Amir Zeldes. 2022.
\newblock \href {https://doi.org/10.5210/dad.2022.102} {Can we fix the scope for coreference? problems and solutions for benchmarks beyond {OntoNotes}}.
\newblock \emph{Dialogue \& Discourse}, 13(1):41--62.

\bibitem[{Zeldes(2024)}]{gum-wiki}
Amir Zeldes. 2024.
\newblock \href {https://wiki.gucorpling.org/gum/entities} {Entity and {I}nformation {S}tatus {A}nnotation}.
\newblock GUM Annotation Wiki.

\bibitem[{Zhang and Zeldes(2017)}]{Zhang2017GitDOXAL}
Shuo Zhang and Amir Zeldes. 2017.
\newblock \href {https://aaai.org/papers/619-flairs-2017-15451/} {{GitDOX}: A linked version controlled online {XML} editor for manuscript transcription}.
\newblock In \emph{Proceedings of the Thirtieth International Florida Artificial Intelligence Research Society Conference (FLAIRS 2017)}, pages 619--623.

\bibitem[{Zhu et~al.(2021)Zhu, Pradhan, and Zeldes}]{zhu-etal-2021-ontogum}
Yilun Zhu, Sameer Pradhan, and Amir Zeldes. 2021.
\newblock \href {https://doi.org/10.18653/v1/2021.acl-short.59} {{O}nto{GUM}: Evaluating contextualized {SOTA} coreference resolution on 12 more genres}.
\newblock In \emph{Proceedings of the 59th Annual Meeting of the Association for Computational Linguistics and the 11th International Joint Conference on Natural Language Processing (Volume 2: Short Papers)}, pages 461--467, Online. Association for Computational Linguistics.

\end{thebibliography}

\section*{Language Resource References}
\label{lr:ref}
\bibliographystylelanguageresource{lrec2026-natbib}
\bibliographylanguageresource{languageresource}

\appendix

\section{LLM Baseline Prompts}
\label{sec:appendix_prompts}

\subsection{Anaphor Recognition}

\begin{quote}

 You are a linguistic analyst whose job is to find cases of bridging anaphora: mentions of newly introduced entities (noun phrases) in a text, for which a reader would need to refer back to a previously mentioned, non-identical entity to resolve their meaning. There are several classes of bridging anaphors, any of which should be identified in the text being analyzed. In the following examples, the bridging anaphor is surrounded by *asterisks*.
    
comparison-relative: The anaphor is preceded by a comparative marker (other, another, same, more, ordinal modifiers, comparative adjectives, superlatives, etc.) which implies a comparison to the antecedent. For example: "The children... *another child*" (=another with comparison to the aforementioned children); similar cases may be *similar children*, *older children* (compared to the aforementioned children), etc.
    
comparison-sense: the semantic type of a phrase requires a previous mention to identify it, for example "the Italian "estaurant... *a Chinese one*" (we can't know "a Chinese one" is a restaurant without referring back to the Italian restaurant), or "*another one*", "*the others*" etc.
    
comparison-time: the anaphor refers to a specific time/timeframe which is understandable with reference to the antecedent, for example: "Tuesday, February 2nd ... *the following week*"
    
entity-meronomy: the anaphor is a subunit of the antecedent (part-whole), including physical subunits, portion-substance relations, and regions/subsections. For example: "the house ... *the door*" (=of the house).
    
entity-associative: the anaphor is an attribute or closely associated entity of the antecedent, including both prototypical and inducible associations: "a wedding ... *the bride*" (=the bride at that wedding), implicit arguments of a predicate or a verbal nominalization: "a play... *the performance*" (=of the play), relational nouns: "a murder ... *the victim*"
    
entity-property: the anaphor is a physical or intangible property of the antecedent (e.g., smell, length, size, style, etc.): "the tea... *the sweet aroma*"
    
entity-resultative: the anaphor is logically inferable from the antecedent (e.g., result, transformation/transmutation, cause): "the dough ... *the bread*" (=the dough becomes bread after baking)
    
set-member: the anaphor is an element of the antecedent set, including groups-member relations and classes-instances: "the cars ... *the Mazda*", additionally indefinite members to definite sets: "a candle on each cupcake... *the candles*"
    
set-subset: the anaphor is a subset of the antecedent set: "the cars ... *the Mazdas*" (not all Mazdas, just the subset among the aforementioned cars)

set-span-interval: the anaphor is a sub-span of a spatial or temporal interval defined by the antecedent: "last week... *Wednesday*" (=Wednesday of last week), "Sunday... *the morning* (=the morning portion of that Sunday)"
    
other: the anaphor requires a previous entity for interpretation, but it doesn't fit into any of the above categories. This is a rare class.
    
There are also some exceptions which should NOT be identified as bridging anaphora:
    
Coreference: If an entity has a previous mention, it cannot be an instance of bridging. For instance, in "Catherine and Henry had their wedding last week. The bride was very beautiful", even though there is an associative relationship between the wedding and the bride, since "the bride" corefers with "Catherine", which has already been introduced to the discourse, "the bride" is not eligible to be an instance of bridging.
    
Bridging-contained: If the entity one would need to refer back to in order to understand the bridging anaphor is a direct modifier in the noun phrase of the potential bridging anaphor, e.g. "the focus of the story" or "two of them", it should not be annotated as bridging. In other words, the previous antecedent entity must be outside of the nominal phrase containing the anaphor. An entity that is followed by a prepositional phrase or a relative clause is sufficiently qualified and is thus NOT an instance of bridging.
    
Generics/Situational bridging: Entities that are accessible due to general world knowledge or situational context are not considered instances of bridging, i.e., if it doesn’t have a previous associated antecedent entity to be bridging from, it cannot be bridging.
    
Possession with an explicit possessive: If the potential bridging anaphor contains an explicit possessive which corefers with the associative antecedent entity, no bridging relation is necessary. Explicit coreference between the associative antecedent and the possessive is sufficient (e.g., [Mark]…[his] house → no bridging, coreference between "Mark" and "his"). Contrast this with [the family] … *the house* → bridging, since we cannot interpret which house it is (the house of the family) without referring to "the family", which is outside of the anaphor phrase.
    
Here are 2 an examples of the task:
    
Please return a list all of the bridging anaphors in the following text in the order in which they appear. Output the anaphor mention phrase exactly as it appears in the text. If there are no bridging anaphors, return an empty list.

Text:

... with their friends to a picnic. The picnic was supposed to take place in a grove, but the shade wasn't enough, so they had to find a different place. Conny started to say ...

Answer(s):

["the shade", "a different place"]
    
Please return a list all of the bridging anaphors in the following text in the order in which they appear. Output the anaphor mention phrase exactly as it appears in the text. If there are no bridging anaphors, return an empty list.

Text:

... making this technique the basis of training for all types of dance . While dancing ballet takes dedication and requires serious training , you can learn the basics to prepare yourself for further study . Learn to get ready for practicing...

Answer(s):

["the basics", "further study"]
    
Please return a list all of the bridging anaphors in the following text in the order in which they appear. Output the anaphor mention phrase exactly as it appears in the text. If there are no bridging anaphors, return an empty list.

Text:

\{text\}

Answer(s):

\end{quote}

\subsection{Antecedent Selection}

\begin{quote}

You are a linguistic analyst whose job is to select the associative antecedent for of a bridging anaphor: mentions of newly introduced entities (noun phrases) in a text, for which a reader would need to refer back to a previously mentioned, non-identical entity (the antecedent) to resolve their meaning. There are several classes of bridging instances, defined by the associative relationship between the bridging anaphor and its associative antecedent. In the following examples, the bridging antecedent is surrounded by *asterisks* and the bridging anaphor is surrounded by \{\{double curly brackets\}\}.

comparison-relative: The anaphor is preceded by a comparative marker (other, another, same, more, ordinal modifiers, comparative adjectives, superlatives, etc.) which implies a comparison to the antecedent. For example: "*The children* ... \{\{another child\}\}" (=another with comparison to the aforementioned children); similar cases may be \{\{similar children\}\}, \{\{older children\}\}(compared to the aforementioned children), etc.

comparison-sense: the semantic type of a phrase requires a previous mention to identify it, for example "*the Italian restaurant* ... \{\{a Chinese one\}\}" (we can't know "a Chinese one" is a restaurant without referring back to the Italian restaurant), or "\{\{another one\}\}", "\{\{the others\}\}" etc.

comparison-time: the anaphor refers to a specific time/timeframe which is understandable with reference to the antecedent, for example: "*Tuesday, February 2nd* ... \{\{the following week\}\}"

entity-meronomy: the anaphor is a subunit of the antecedent (part-whole), including physical subunits, portion-substance relations, and regions/subsections. For example: "*the house* ... \{\{the door\}\}" (=of the house).

entity-associative: the anaphor is an attribute or closely associated entity of the antecedent, including both prototypical and inducible associations: "*a wedding* ... \{\{the bride\}\}" (=the bride at that wedding), implicit arguments of a predicate or a verbal nominalization: "*a play*... \{\{the performance\}\}" (=of the play), relational nouns: "*a murder* ... \{\{the victim\}\}"

entity-property: the anaphor is a physical or intangible property of the antecedent (e.g., smell, length, size, style, etc.): "*the tea* ... \{\{the sweet aroma\}\}"

entity-resultative: the anaphor is logically inferable from the antecedent (e.g., result, transformation/transmutation, cause): "*the dough* ... \{\{the bread\}\}" (=the dough becomes bread after baking)

set-member: the anaphor is an element of the antecedent set, including groups-member relations and classes-instances: "*the cars* ... \{\{the Mazda\}\}", additionally indefinite members to definite sets: "*a candle* on each cupcake ... \{\{the candles\}\}"

set-subset: the anaphor is a subset of the antecedent set: "*the cars* ... \{\{the Mazdas\}\}" (not all Mazdas, just the subset among the aforementioned cars)

set-span-interval: the anaphor is a sub-span of a spatial or temporal interval defined by the antecedent: "*last week* ... \{\{Wednesday\}\}" (=Wednesday of last week), "*Sunday* ... \{\{the morning\}\}" (=the morning portion of that Sunday)

other: the anaphor requires a previous entity for interpretation, but it doesn't fit into any of the above categories. This is a rare class.

Here are 2 an examples of the task:

Please return a single string for associative antecedent of the bridging anaphor surrounded by \{\{double curly brackets\}\}. Output the antecedent mention phrase exactly as it appears in the text. If there is no associative antecedent, return "no antecedent". The antecedent you are returning CANNOT be the same as the bracketed anaphor.

Text:

... with their friends to a picnic. The picnic was supposed to take place in a grove, but \{\{the shade\}\} wasn't enough, so they had to find a different place. Conny started to say ...

Answer:

a grove

Please return a single string for associative antecedent of the bridging anaphor surrounded by \{\{double curly brackets\}\}. Output the antecedent mention phrase exactly as it appears in the text. If there is no associative antecedent, return "no antecedent".

Text:

... making this technique the basis of training for all types of dance . While dancing ballet takes dedication and requires serious training , you can learn the basics to prepare yourself for \{\{further study\}\} . Learn to get ready for practicing ...

Answer:

ballet

Please return a single string for associative antecedent of the bridging anaphor surrounded by \{\{double curly brackets\}\}. Output the antecedent mention phrase exactly as it appears in the text. If there is no associative antecedent, return "no antecedent".

Text:

\{text\}

Answer:
    
\end{quote}

\subsection{Subtype Categorization}

\begin{quote}

You are a linguistic analyst whose job is to select the subtype classification for a bridging anaphor - antecedent pair: mentions of newly introduced entities (the anaphor) in a text, for which a reader would need to refer back to a previously mentioned, non-identical entity (the antecedent) to resolve their meaning. There are several classes of bridging instances, defined by the associative relationship between the bridging anaphor and its associative antecedent. In the following subtype examples, the bridging antecedent is surrounded by *asterisks* and the bridging anaphor is surrounded by \{\{double curly brackets\}\}.

comparison-relative: The anaphor is preceded by a comparative marker (other, another, same, more, ordinal modifiers, comparative adjectives, superlatives, etc.) which implies a comparison to the antecedent. For example: "*The children* ... \{\{another child\}\}" (=another with comparison to the aforementioned children); similar cases may be \{\{similar children\}\}, \{\{older children\}\} (compared to the aforementioned children), etc.

comparison-sense: the semantic type of a phrase requires a previous mention to identify it, for example "*the Italian restaurant* ... \{\{a Chinese one\}\}" (we can't know "a Chinese one" is a restaurant without referring back to the Italian restaurant), or "\{\{another one\}\}", "\{\{the others\}\}" etc.

comparison-time: the anaphor refers to a specific time/timeframe which is understandable with reference to the antecedent, for example: "*Tuesday, February 2nd* ... \{\{the following week\}\}"

entity-meronomy: the anaphor is a subunit of the antecedent (part-whole), including physical subunits, portion-substance relations, and regions/subsections. For example: "*the house* ... \{\{the door\}\}" (=of the house).

entity-associative: the anaphor is an attribute or closely associated entity of the antecedent, including both prototypical and inducible associations: "*a wedding* ... \{\{the bride\}\}" (=the bride at that wedding), implicit arguments of a predicate or a verbal nominalization: "*a play*... \{\{the performance\}\}" (=of the play), relational nouns: "*a murder* ... \{\{the victim\}\}"

entity-property: the anaphor is a physical or intangible property of the antecedent (e.g., smell, length, size, style, etc.): "*the tea* ... \{\{the sweet aroma\}\}"

entity-resultative: the anaphor is logically inferable from the antecedent (e.g., result, transformation/transmutation, cause): "*the dough* ... \{\{the bread\}\}" (=the dough becomes bread after baking)

set-member: the anaphor is an element of the antecedent set, including groups-member relations and classes-instances: "*the cars* ... \{\{the Mazda\}\}", additionally indefinite members to definite sets: "*a candle* on each cupcake ... \{\{the candles\}\}"

set-subset: the anaphor is a subset of the antecedent set: "*the cars* ... \{\{the Mazdas\}\}" (not all Mazdas, just the subset among the aforementioned cars)

set-span-interval: the anaphor is a sub-span of a spatial or temporal interval defined by the antecedent: "*last week* ... \{\{Wednesday\}\}" (=Wednesday of last week), "*Sunday* ... \{\{the morning\}\}" (=the morning portion of that Sunday)

other: the anaphor requires a previous entity for interpretation, but it doesn't fit into any of the above categories. This is a rare class.

Here are 2 an examples of the task:
    
In the following text, a bridging anaphora is marked with \{\{double curly brackets\}\} and the corresponding antecedent is surrounded by *asterisks*. Read the following text and for the bridging anaphor-antecedent pair, classify the variety of bridging subtype relation (defined above) that holds between the two entities. Multiple subtypes may apply to a single pair. Output a string of all applicable subtypes, connected by semicolons (no spaces).

The possible subtype labels are as follows:

comparison-relative

comparison-sense

comparison-time

entity-associative

entity-meronomy

entity-property

entity-resultative

set-member

set-subset

set-span-interval

other

Antecedent Text:

... with their friends to a picnic. The picnic was supposed to take place in *a grove*, but the shade wasn't enough, so they had to find a different place. Conny started to say ...

Anaphor Text:

... to a picnic. The picnic was supposed to take place in a grove, but \{\{the shade\}\} wasn't enough, so they had to find a different place. Conny started to say ...

Answer:

entity-associative

In the following text, a bridging anaphora is marked with \{\{double curly brackets\}\} and the corresponding antecedent is surrounded by *asterisks*. Read the following text and for the bridging anaphor-antecedent pair, classify the variety of bridging subtype relation (defined above) that holds between the two entities. Multiple subtypes may apply to a single pair. Output a string of all applicable subtypes, connected by semicolons (no spaces).

The possible subtype labels are as follows:

comparison-relative

comparison-sense

comparison-time

entity-associative

entity-meronomy

entity-property

entity-resultative

set-member

set-subset

set-span-interval

other

Antecedent Text:

... this technique the basis of training for all types of dance . While dancing *ballet* takes dedication and requires serious training , you can learn the basics to prepare yourself for further study . Learn to get ready for practicing ...

Anaphor Text:

... making this technique the basis of training for all types of dance . While dancing ballet takes dedication and requires serious training , you can learn the basics to prepare yourself for \{\{further study\}\} . Learn to get ready for ...

Answer:

comparison-relative

In the following text, a bridging anaphora is marked with \{\{double curly brackets\}\} and the corresponding antecedent is surrounded by *asterisks*. Read the following text and for the bridging anaphor-antecedent pair, classify the variety of bridging subtype relation (defined above) that holds between the two entities. Multiple subtypes may apply to a single pair. Output a string of all applicable subtypes, connected by semicolons (no spaces).

The possible subtype labels are as follows:

comparison-relative

comparison-sense

comparison-time

entity-associative

entity-meronomy

entity-property

entity-resultative

set-member

set-subset

set-span-interval

other

Antecedent Text:

... \{antecedent\_text\} ...

Anaphor Text:

... \{anaphor\_text\} ...

Answer:
    
\end{quote}

\end{document}